\documentclass[letterpaper]{article} % DO NOT CHANGE THIS
\usepackage{aaai2027}  % DO NOT CHANGE THIS
\nocopyright
\usepackage[hyphens]{url}  % DO NOT CHANGE THIS
\usepackage{graphicx} % DO NOT CHANGE THIS
\usepackage{natbib}  % DO NOT CHANGE THIS AND DO NOT ADD ANY OPTIONS TO IT
\usepackage{caption} % DO NOT CHANGE THIS AND DO NOT ADD ANY OPTIONS TO IT
\usepackage{algorithm}
\usepackage{algorithmic}
\usepackage{amsmath}
\usepackage{amssymb}
\usepackage{tikz}
\usepackage{newfloat}
\usepackage{listings}
\DeclareCaptionStyle{ruled}{labelfont=normalfont,labelsep=colon,strut=off} % DO NOT CHANGE THIS
\floatstyle{ruled}
\newfloat{listing}{tb}{lst}{}
\floatname{listing}{Listing}

\usepackage{booktabs}

\title{From Relevance to Execution Utility: Reward-Aware\\ Dynamic Execution Gating for Skill-Based LLM Agents}
\author{
Liang He$^{1}$ \quad Jingbo Wen$^{2}$ \quad Hongyu Gu$^{3}$ \quad Hao Li$^{3}$ \\[2pt]
Haoyu Wang$^{4}$ \quad Yixiong Chen$^{5}$ \quad Kangning Cui$^{6}$ \quad Xilu Wang$^{7,\ast}$ \\[4pt]
$^{1}$Tongji University \quad
$^{2}$The University of Sydney \quad
$^{3}$University of Science and Technology of China \\[2pt]
$^{4}$Nankai University \quad
$^{5}$Johns Hopkins University \quad
$^{6}$City University of Hong Kong \quad
$^{7}$University of Surrey \\[4pt]
\texttt{wangxilu@surrey.ac.uk} \\[2pt]
$^\ast$Corresponding author
}
\affiliations{
    \textsuperscript{\rm }
}

\begin{document}

\maketitle

\begin{abstract}
Agent skills are increasingly used to equip large language model (LLM) agents with reusable procedural knowledge. Although recent work has substantially improved skill retrieval due to the increasing skill libraries, retrieving a plausible skill bundle does not guarantee that executing it is worthwhile. Since every skill-conditioned rollout is computationally expensive, deciding whether a retrieved bundle should be executed has become an increasingly important challenge. To this end, we introduce the \textbf{Reward-Aware Dynamic Execution Gate (RADEG)}, a lightweight, retriever-agnostic decision layer between skill retrieval and agent execution. RADEG learns a low-cost surrogate model that predicts the execution utility of a query--bundle pair before the expensive rollout is launched. To obtain informative supervision while controlling for task difficulty, we locally perturb each retrieved bundle by deleting, adding, or replacing one skill, producing matched same-query rollouts that isolate the effect of bundle composition on verifier reward. During deployment, RADEG updates only a warm-started logistic head as new verifier feedback becomes available, enabling inexpensive adaptation of the execute/skip boundary without retraining either the retriever or the agent.
Under a query-level held-out evaluation on 288 collected rollouts, RADEG substantially reduces unnecessary agent executions while preserving a large fraction of the downstream verifier reward. It consistently outperforms relevance-based and random gating across different execution budgets, demonstrating that execution-aware surrogate modeling provides a practical and cost-effective complement to skill retrieval.
\end{abstract}

% Uncomment the following to link to your code, datasets, an extended version or similar.
% You must keep this block between (not within) the abstract and the main body of the paper.
% Make sure that you do not de-anonymize yourself with these links.
% \begin{links}
%     \link{Code}{https://aaai.org/example/code}
%     \link{Datasets}{https://aaai.org/example/datasets}
%     \link{Extended version}{https://aaai.org/example/extended-version}
% \end{links}

\section{Introduction}
Agent skills have emerged as a promising mechanism for enabling LLM-based agents to tackle complex, long-horizon tasks \citep{wang2024voyager,li2026skillsbench, anand2026aeval}. A skill encapsulates reusable procedural knowledge in a modular package of instructions, executable code, and auxiliary resources that can be loaded on demand at inference time \citep{xu2026agentskills,mi2026skillpro,kevin2026evaluating}. Since a skill library may far exceed the agent's available context, each incoming query requires a retrieval decision: which small bundle of skills should be supplied to the agent for the task at hand?
 
Existing skill-augmented agents commonly treat this decision as a retrieval or routing problem \citep{yang2026skillseek, gao2026skillreducer}. Early systems such as Voyager retrieve skills from a continually expanding library by embedding similarity \citep{wang2024voyager}. More recent systems operate over substantially larger repositories: SkillFlow \citep{li2025skillflow} progressively narrows the candidate set through dense retrieval, two rounds of cross-encoder reranking, and LLM-based selection, while SkillRouter \citep{zheng2026skillrouter} pairs a compact bi-encoder retriever with a cross-encoder reranker that reads the full skill body. A further line of work moves beyond item-level relevance by explicitly modelling relationships among skills, organizing them into capability trees and DAG-based execution pipelines \citep{li2026agentskillos}, dependency graphs \citep{liu2026graphofskills}, hierarchical and role-structured groups \citep{zeng2026groupofskills}, or typed directed graphs \citep{bai2026skilldag}. These structures allow the retriever to return bounded bundles whose members complement one another. The retrieved skills are then compiled into the agent's context, arranged into an executable workflow. 
 
Despite this progress, skill selection is still predominantly optimized through item-level relevance scores or predefined structural relationships. They generally ignore the end-to-end utility of the selected bundle for a specific task and agent, resulting in expensive execution cost. Evaluation practice mirrors this focus: skill retrievers and routing benchmarks are typically scored with ranking-oriented metrics such as Hit@$K$, Recall@$K$, MRR, and NDCG \citep{li2025skillflow,zheng2026skillrouter}. Such metrics measure whether annotated skills are ranked highly, but not whether executing the target agent with the retrieved bundle will actually produce a useful outcome. Consistent with this concern, prior work reports skill shadowing as libraries grow \citep{song2026skillshadowing}, negative transfer across agents \citep{huang2026skilllens}, and failures to correctly incorporate even successfully retrieved gold skills \citep{su2026sra}.
 
We refer to this mismatch as the \emph{relevance--utility gap}. Retrieval
answers which skills appear relevant to a query. However, it does not necessarily
answer whether the resulting query--bundle pair is worth executing. This gap
is particularly consequential because agent execution is typically far more
expensive than retrieval: invoking the underlying agent consumes API calls,
tokens, tool interactions, and wall-clock time, even when the final verifier
reward is zero. A retrieval-centric pipeline therefore risks spending a
substantial fraction of its execution budget on bundles that appear relevant
but do not yield downstream utility.

%We examine this gap using 72 tasks from SkillsBench \citep{li2026skillsbench}. For each bundle selected by Graph-of-Skills (GoS), we construct three controlled perturbations: removing the skill with the highest personalized PageRank (PPR) score, adding a semantically irrelevant skill, and replacing a selected skill with a semantically similar alternative. Accordingly, 22 of the 72 tasks (30.6\%) exhibit a reward change under at least one perturbation. Moreover, aggregate PPR relevance scores are poor predictors of positive verifier reward, with AUROC values between 0.418 and 0.435. Relevance remains necessary for identifying plausible candidate skills, but these results show that it is not a reliable proxy for downstream execution utility.

We empirically examine this relevance--utility gap on 72 SkillsBench tasks
\citep{li2026skillsbench}. Starting from bundles retrieved by
Graph-of-Skills (GoS) \citep{liu2026graphofskills}, we construct local variants by deleting, adding, or replacing
one skill. We find that 22 of the 72 tasks change reward under at least one variant, while aggregate GoS relevance scores provide little signal for
positive verifier reward (detailed in Motivating Study). 

These findings motivate a
separate post-retrieval decision that predicts whether a retrieved
query--bundle pair is worth executing. To address this problem, we propose the \textbf{Reward-Aware Dynamic Execution Gate (RADEG)}, a lightweight post-retrieval, pre-execution module that estimates whether a retrieved query--bundle pair is likely to produce non-zero verifier reward. RADEG learns from previously observed $(\text{query}, \text{bundle}, \text{reward})$ rollouts and makes its execution decision before the underlying agent is invoked, leaving both the retriever and the agent unchanged. Therefore, RADEG can serve as a retrieval-agnostic decision layer on top of existing skill-augmented pipelines. Moreover, when new execution feedback becomes available, the gate can update
its decision boundary without retraining or modifying the agent itself. Our main contributions are as follows:
\begin{enumerate}
\item We provide a controlled empirical diagnosis of the relevance--utility gap in multi-skill retrieval. Our perturbation analysis shows that downstream reward is sensitive to bundle composition, while relevance scores provide little predictive signal for positive execution reward.
 
\item To improve LLM agents' efficiency, we formulate post-retrieval execution gating for skill-based LLM agents and introduce RADEG, a lightweight retrieval-agnostic module that uses observed execution feedback to estimate a query--bundle pair's utility before invoking the agent.

\item We evaluate RADEG on 288 logged rollouts using a query-level held-out protocol. At its default decision rule, RADEG skips 68\% of agent calls while retaining 61\% of the total reward. Under a fixed 20\% call budget, it retains 40\% of the reward, substantially outperforming a relevance score-based and random execution.
\end{enumerate}

\section{Related Work}

\paragraph{Skill retrieval and bundle construction.}
As skill libraries grow beyond the agent's context budget, recent work has
treated skill access as a retrieval and routing problem. Systems such as
SkillFlow and SkillRouter combine dense retrieval with progressively more
expensive reranking or selection stages
\citep{li2025skillflow,zheng2026skillrouter}, while SkillRet provides a
large-scale benchmark for evaluating retrieval over realistic skill
libraries \citep{cho2026skillret}. Other work moves beyond independent
query--skill relevance by modelling compatibility and ambiguity among skills
retrieved together
\citep{wang2026skillnotdocument,ding2026skillresolve}. A complementary line of research explicitly models multi-skill structure
through capability hierarchies, dependency graphs, role-structured groups,
or typed relations
\citep{li2026agentskillos,liu2026graphofskills,zeng2026groupofskills,bai2026skilldag}.
These structures support the construction of compact bundles subject to
dependencies and deployment constraints
\citep{zheng2026skillselect}. RADEG is complementary to these methods:
rather than constructing or reranking a bundle, it evaluates whether a
retrieved query--bundle pair is likely to yield downstream execution reward.

\paragraph{From successful retrieval to successful execution.}
Recent work shows that retrieving an appropriate skill does not guarantee successful downstream execution. SRA-Bench separates skill augmentation into retrieval, incorporation, and application, demonstrating that agents may fail
to load or correctly use even a gold skill \citep{su2026sra}. Related studies
identify execution failures caused by planning drift, verifier mismatch, skill shadowing, and negative transfer across agents \citep{liu2026graphofskills,song2026skillshadowing,huang2026skilllens}.
SkillsBench further shows that the benefit of curated skills varies across
tasks and skill configurations \citep{li2026skillsbench}. These findings
motivate RADEG's post-retrieval execution gate: rather than modifying the
retrieved bundle, it predicts from verifier feedback whether executing the
complete query--bundle pair is likely to yield non-zero reward.

\paragraph{Cost-aware routing and selective execution.}
Cost-aware LLM systems avoid low-value computation by routing queries across
models or selectively using retrieved support. FrugalGPT and RouteLLM allocate
queries among LLMs to balance quality and cost, while recent memory- and
retrieval-control methods decide whether retrieved guidance should influence
an agent or whether a failed retrieval should trigger corrective skills
\citep{chen2024frugalgpt,ong2025routellm,iscan2026learningwhen,
wei2026skillrag}. RADEG operates
after skill retrieval but before agent execution by predicting whether the skill bundle should be evaluated.

\begin{figure*}[t]
    \centering
    \includegraphics[width=1\textwidth]{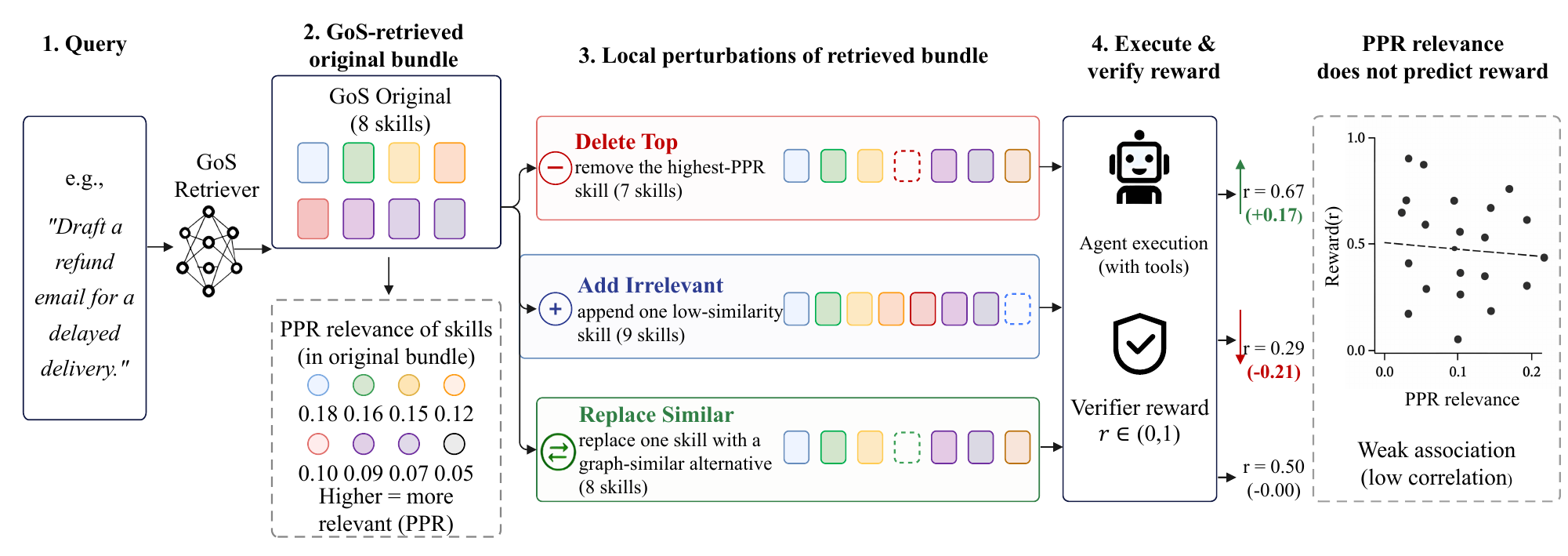}
    \caption{
    Overview of the motivating study. For each query, GoS retrieves an eight-skill bundle ranked by PPR relevance. We construct three local variants and observe that each perturbation can increase, decrease, or leave
    verifier reward unchanged, showing that
    higher retrieval relevance does not reliably imply higher execution
    utility.
    }
    \label{fig:motivating_study}
\end{figure*}
\section{Motivating Study}
\label{sec:motivation}
A relevance-based retriever maps a query $q$ to a skill bundle
$b=\mathcal{R}(q)$, which is then supplied to an agent and evaluated by a
task-specific verifier with reward $r(q,b)\in[0,1]$. We investigate whether
retrieval relevance is also a reliable proxy for downstream execution
utility. We use 72 tasks from SkillsBench \citep{li2026skillsbench}. For each task, GoS retrieves an eight-skill bundle
\citep{liu2026graphofskills}. Starting from this bundle, we construct three
local variants: \textbf{Delete Top}, which removes the highest-PPR skill; \textbf{Add Irrelevant}, which appends a low-similarity skill; and \textbf{Replace Similar}, which replaces one skill with a graph-similar alternative. Together with the original GoS bundle, this produces $72\times4=288$ query--bundle pairs.
Each pair is executed once under the same agent and verifier configuration. Figure~\ref{fig:motivating_study} summarizes the perturbation procedure and resulting task-level outcomes.
%Because agent execution is stochastic, we interpret the experiment as a matched analysis of sensitivity to bundle composition rather than as a causal estimate of an individual skill edit.
\paragraph{Bundle composition affects realized utility.}
Table~\ref{tab:bundle_effects} shows that no perturbation is uniformly
beneficial: each improves some tasks and degrades others. Overall, 22 of the
72 tasks exhibit a reward change under at least one perturbation. Across the
216 original--perturbed comparisons, we observe 17 zero-to-positive and 15
positive-to-zero transitions. At least one tested alternative outperforms the
original GoS bundle on 13 tasks, including 11 tasks that change from zero to
positive reward. Thus, a bundle selected according to retrieval relevance
does not necessarily provide the highest observed execution utility.

\begin{table}[t]
\centering
\scriptsize
\setlength{\tabcolsep}{4pt}
\begin{tabular}{lcccc}
\toprule
Condition
& Mean $r$
& Success ($r>0$)
& \multicolumn{1}{c}{Reward vs.\ GoS} & \multicolumn{1}{c}{Status flips} \\
& & & $\uparrow\,/\,=\,/\,\downarrow$ & $0\!\to\!+\,/\,+\!\to\!0$ \\
\midrule
GoS Original & 0.285 & 23 (31.9\%) & -- & -- \\
Delete Top & 0.304 & 25 (34.7\%) & 8 / 57 / 7 & 7 / 5 \\
Add Irrelevant & 0.269 & 22 (30.6\%) & 6 / 60 / 6 & 4 / 5 \\
Replace Similar & 0.295 & 24 (33.3\%) & 8 / 58 / 6 & 6 / 5 \\
\midrule
\textit{Best observed} & 0.433 & 34 (47.2\%) & 13 / 59 / 0 & 11 / 0 \\
\bottomrule
\end{tabular}
\caption{
Task-level comparison of each bundle condition with GoS Original.
$\uparrow$, $=$, and $\downarrow$ denote numbers of tasks with higher,
equal, and lower verifier reward than GoS. The last row reports the
best observed reward and serves only as a post-hoc upper bound.
}
\label{tab:bundle_effects}
\end{table}

\paragraph{PPR relevance provides little utility signal.}
GoS assigns each skill $s$ a query-dependent Personalized PageRank (PPR)
score $\pi_q(s)$ by propagating the initial semantic and lexical relevance
of query $q$ through the skill dependency graph \citep{liu2026graphofskills}. Thus,
$\pi_q(s)$ measures how relevant a skill appears to the query under the GoS
retrieval model. For a query--bundle pair $(q,b)$, let $\mathcal{S}^{\mathrm{PPR}}_b$ denote
the skills in $b$ that receive PPR scores from GoS. We summarize the
retrieval relevance of the bundle using
\[
    R_{\mathrm{sum}}(q,b)
    =
    \sum_{s\in\mathcal{S}^{\mathrm{PPR}}_b}\pi_q(s),
    \qquad
    R_{\mathrm{max}}(q,b)
    =
    \max_{s\in\mathcal{S}^{\mathrm{PPR}}_b}\pi_q(s).
\]
Here, $R_{\mathrm{sum}}$ measures the total propagated relevance assigned to the GoS-scored portion of the bundle, while $R_{\mathrm{max}}$ captures the
relevance of its highest-ranked skill. We measure their association with continuous verifier reward using Spearman
correlation and their ability to rank positive-reward executions using AUROC. As shown in Table~\ref{tab:ppr_gap}, both correlations are close to
zero, while the AUROCs are 0.435 and 0.418. These results provide no evidence
that higher aggregate PPR relevance reliably identifies query--bundle pairs
that will obtain positive reward.
%For a skill added after GoS retrieval, such as the injected skill in Add Irrelevant, the statistics are computed over the original GoS-scored skills. Because deleting a skill also changes the number of scored bundle members, $R_{\mathrm{sum}}$ may partially reflect bundle size; $R_{\mathrm{max}}$ therefore provides a complementary measure that does not directly accumulate scores across skills.

\begin{table}[t]
\centering
\small
\begin{tabular}{lc}
\toprule
Metric & Value \\
\midrule
Spearman($R_{\mathrm{sum}}$, $r$)
    & $-0.089$ \\
Spearman($R_{\mathrm{max}}$, $r$)
    & $-0.131$ \\
AUROC($R_{\mathrm{sum}}\rightarrow r>0$)
    & 0.435 \\
AUROC($R_{\mathrm{max}}\rightarrow r>0$)
    & 0.418 \\
\bottomrule
\end{tabular}
\caption{Predictive value of aggregate GoS relevance scores for downstream
execution reward over 288 rollouts.}
\label{tab:ppr_gap}
\end{table}
Together, these observations reveal a relevance--utility gap: retrieval can identify a plausible bundle without determining whether executing it is worth the cost. This motivates RADEG, a lightweight post-retrieval gate that predicts execution utility before launching the agent.
Additional analyses of PPR relevance are reported in the Technical Supplement Section 2.4 (PPR permutation test).

\section{Reward-Aware Dynamic Execution Gate}
\label{sec:method}

\begin{figure*}[t]
    \centering
    \includegraphics[width=1\textwidth]
    {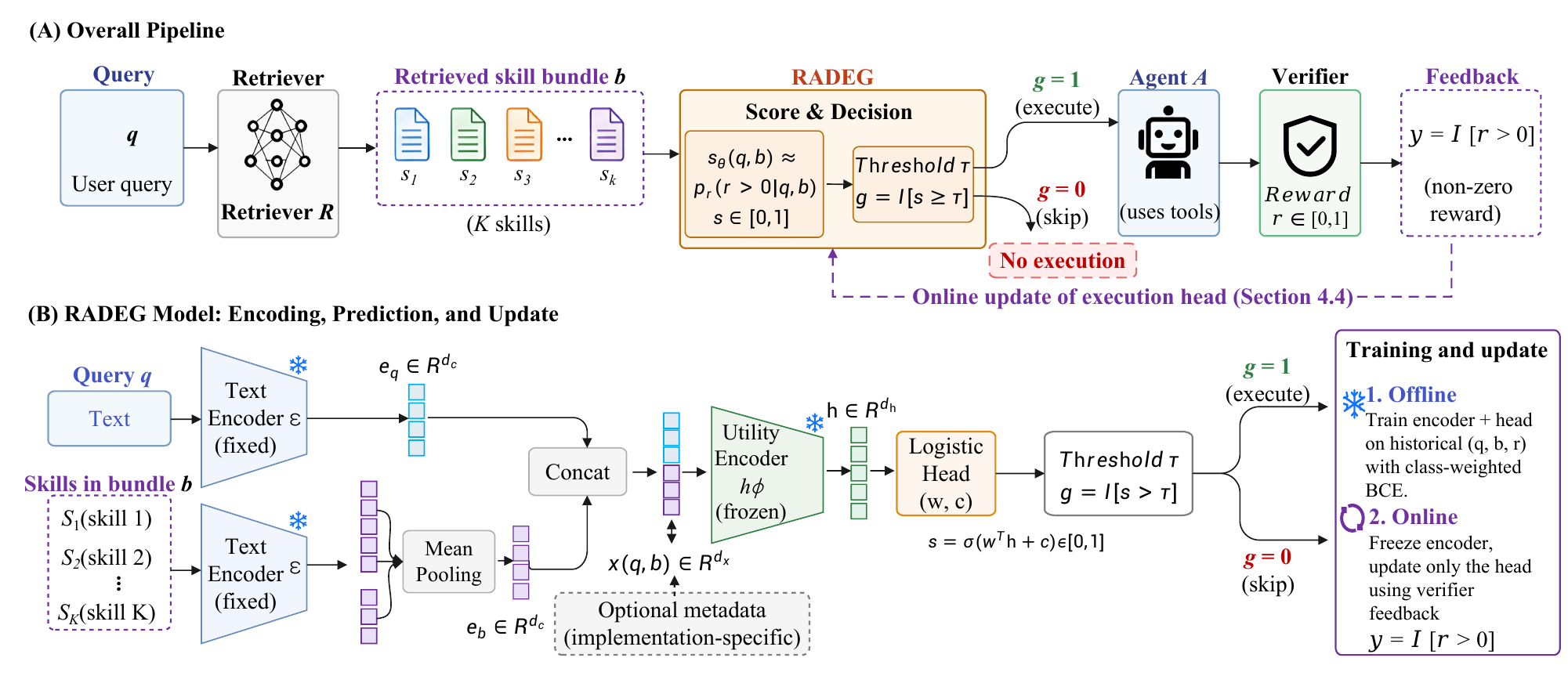}
    \caption{
    Overview of RADEG. Given a query, the upstream retriever first constructs
    a candidate skill bundle. RADEG encodes the query and the retrieved skills into a joint query--bundle representation and predicts the probability that executing the downstream agent will obtain non-zero
    verifier reward. During deployment, the encoder
    is frozen and only the lightweight execution head is updated using newly
    observed verifier feedback.
    }
    \label{fig:radeg_overview}
\end{figure*}

RADEG is a post-retrieval decision module that determines whether a retrieved
query--bundle pair is worth executing. As illustrated in
Figure~\ref{fig:radeg_overview}, it complements, rather than replaces, the
upstream retriever: the retriever identifies a relevant bundle, whereas
RADEG estimates whether invoking the downstream agent with that bundle is likely to produce verifiable utility. Both the retriever and the agent remain
unchanged.
\subsection{Execution-Gating Objective}
\label{sec:problem_formulation}

Let $\mathcal{R}$ be an upstream skill retriever. Given a user query $q$, it
returns a skill bundle $b=\mathcal{R}(q)$.
A downstream agent $\mathcal{A}$ executes the task using $b$, and a
task-specific verifier $\mathcal{V}$ returns $r(q,b) = \mathcal{V}\!\left(\mathcal{A}(q,b)\right)
    \in[0,1]$.

We consider an execution useful when it obtains any non-zero verifier reward $y(q,b) = \mathbb{I}[r(q,b)>0]$. This definition includes partial rewards. In SkillsBench, $0<r<1$ indicates
that the execution passes a non-empty subset of the deterministic verifier
tests and therefore produces measurable downstream value.

Given historical execution records
$\mathcal{D} = \{(q_i,b_i,r_i)\}_{i=1}^{N}$, RADEG learns a utility score $s_{\theta}(q,b)
    \approx
    \Pr\!\left(r(q,b)>0\mid q,b\right)$. At inference time, it applies the decision rule
\[
    g_{\theta}(q,b;\tau)
    =
    \mathbb{I}\!\left[s_{\theta}(q,b)\geq\tau\right],
\]
where $\tau$ is an execution threshold. The agent is invoked when
$g_{\theta}=1$ and skipped otherwise. Varying $\tau$ controls the trade-off
between execution frequency and retained verifier reward.

\subsection{Query--Bundle Encoding}
\label{sec:representation}

RADEG estimates utility from the semantic relationship between the query and
the procedural information contained in the retrieved bundle. Let $E:\mathcal{T}\rightarrow\mathbb{R}^{d}$ denote a fixed text encoder. The query representation is  $ e_q=E(q)$.

For each skill $s\in b$, let $d_s$ denote its textual specification. We use
the instructions and description in \texttt{SKILL.md} when available and
otherwise use the available skill metadata. Each skill is encoded
independently, and the bundle representation is obtained by mean pooling:
\[
    e_b
    =
    \frac{1}{|b|}
    \sum_{s\in b}E(d_s).
\]
Mean pooling produces a fixed-dimensional summary of the procedural content
supplied to the agent, independent of the number and ordering of skills in
the bundle.

The core query--bundle representation is
\[
    x_{\mathrm{sem}}(q,b)
    =
    [\,e_q\;\|\;e_b\,],
\]
where $\|$ denotes vector concatenation. This joint representation enables
the utility predictor to learn whether the retrieved procedural content is
appropriate for the given query.

Our evaluated implementation augments this semantic representation with a
small set of features available from the bundle-construction process. These
features are implementation-specific rather than required by the RADEG
formulation and are detailed in Section~\ref{sec:setup}. We denote the
complete model input by $x(q,b)$.

\subsection{Offline Utility Learning}
\label{sec:offline_learning}

RADEG first learns a utility representation from the historical execution
records in $\mathcal{D}$. A neural encoder
$h_{\phi}:\mathbb{R}^{d_x}\rightarrow\mathbb{R}^{d_h}$ maps the input representation to $h_i=h_{\phi}\!\left(x(q_i,b_i)\right)$.
A logistic execution head then estimates the probability of obtaining
non-zero verifier reward:
\[
    s_i
    =
    \sigma\!\left(w^{\top}h_i+c\right),
\]
where $\sigma(\cdot)$ is the sigmoid function. Thus,
\[
    s_i
    \approx
    \Pr(r_i>0\mid q_i,b_i).
\]

The encoder and execution head are jointly trained using class-weighted
binary cross-entropy:
\[
    \mathcal{L}_{\mathrm{off}}
    =
    -\frac{1}{N}
    \sum_{i=1}^{N}
    \left[
        \omega_{+}y_i\log s_i
        +
        (1-y_i)\log(1-s_i)
    \right],
\]
where $y_i=\mathbb{I}[r_i>0]$, and the positive-class weight $\omega_{+} = \frac{N_{\mathrm{neg}}}{N_{\mathrm{pos}}}$ compensates for the imbalance between
zero- and positive-reward executions.

This stage produces a utility encoder $h_{\phi}$ and an initial execution
head $(w,c)$. The encoder captures reusable patterns in the relationship
between queries, skill bundles, and execution outcomes. The learned head
provides the initial decision boundary for dynamic gating.

\subsection{Reward-Aware Dynamic Gating}
\label{sec:dynamic_gate}

After offline training, RADEG freezes the utility encoder $h_{\phi}$ and
updates only the lightweight logistic execution head. This separation
preserves a stable query--bundle representation while allowing the execution
decision boundary to adapt efficiently as new verifier feedback becomes
available.

For the $t$-th query--bundle pair $(q_t,b_t)$, RADEG computes $h_t = h_{\phi}\!\left(x(q_t,b_t)\right)$
and predicts $s_t =\sigma\!\left(w_t^{\top}h_t+c_t\right)$. Based on the prediction, the agent is executed according to $g_t=\mathbb{I}[s_t\geq\tau]$.

The dynamic head is initialized from the output layer learned during offline
training. Whenever verifier feedback $r_t$ is observed, RADEG forms the label $y_t=\mathbb{I}[r_t>0]$
and performs one online gradient update:
\[
    (w_{t+1},c_{t+1})
    =
    (w_t,c_t)
    -
    \eta
    \nabla_{w,c}
    \ell\!\left(
        \sigma(w_t^{\top}h_t+c_t),
        y_t
    \right),
\]
where $\ell$ denotes binary cross-entropy and $\eta$ is the online learning
rate. Only the execution head is updated; the encoder, upstream retriever,
and downstream agent remain fixed. The protocol used to
simulate this feedback process is described in
Section~\ref{sec:setup}.

% For each incoming query, the complete procedure is therefore
% \[
%     q_t
%     \xrightarrow{\mathcal{R}}
%     b_t
%     \xrightarrow{\mathrm{RADEG}}
%     g_t
%     \in\{\textsc{Execute},\textsc{Skip}\}.
% \]
% When execution is selected, the resulting verifier reward can be used to
% update the gate before processing the next query. 

\section{Experiments}
\label{sec:experiments}
%We evaluate RADEG from both predictive and operational perspectives. We first measure utility prediction on unseen queries using leakage-free query-level splits, and then examine reward retention under fixed execution budgets. We further test whether the learned query--bundle utility signal transfers to new downstream agents and can be adapted using lightweight online updates. Finally, we analyze RADEG under selective feedback and study the contribution of its input features.
\begin{table*}[t!]
\centering
\small
\setlength{\tabcolsep}{5pt}
\begin{tabular}{lccccc}
\toprule
Method
& AUROC $\uparrow$
& AUPRC $\uparrow$
& Calls saved $\uparrow$
& Reward retained $\uparrow$
& Reward/call $\uparrow$ \\
\midrule
Always-Execute
& -- & -- & 0 & 100 & 0.309 \\

Random-Gate
& 0.500 & 0.345 & 71 & 30 & 0.313 \\

PPR-Gate
& 0.447 & 0.335 & 46 & 51 & 0.292 \\

Static Head
& 0.517 & 0.375 & 73 & 27 & 0.308 \\

Memory-Corrected Head
& 0.608 & 0.440 & 73 & 28 & 0.321 \\

\textbf{RADEG}
& \textbf{0.717}
& \textbf{0.570}
& 68
& \textbf{61}
& \textbf{0.483} \\

Oracle
& 1.000 & 1.000 & 65 & 100 & 0.892 \\
\bottomrule
\end{tabular}
\caption{
Utility prediction and execution efficiency under the same five query-level
splits. Calls saved and reward retained are reported as percentages.
Operational metrics use $\tau=0.5$.
}
\label{tab:main_comparison}
\end{table*}
\subsection{Experimental Setup}
\label{sec:setup}

\paragraph{Benchmark, data, and execution environment.}
We evaluate RADEG on SkillsBench \citep{li2026skillsbench}. 
The source dataset contains 288 rollouts from 72 queries, with 194 zero-reward and 94 positive-reward executions.
To avoid leakage, all bundle variants of a query are kept in the same partition. We therefore partition by query, using approximately 70\% of queries for training and the remainder as an unseen evaluation stream. The principal comparison uses the same five random query-level splits for every method. A separate 30-split analysis evaluates the uncertainty of RADEG itself. The source rollouts are generated in a Docker-based, Claude-Code-style environment configured with the model identifier Claude Sonnet~4.6 through a third-party API relay. Each rollout is scored by the deterministic, task-specific SkillsBench verifier, which returns a normalized weighted test-pass reward in $[0,1]$.
Additional implementation details and experimental environment are reported in Technical Supplement Section 1.1 (Experimental Environment) and Section 1.2 (Compared Methods and Evaluation Protocol); dataset statistics and perturbation analyses are reported in Section 4.1 (Dataset Composition and Perturbation Effects); end-to-end runtime comparisons and execution-gate overhead are reported in Section 5 (Runtime Analysis).

\paragraph{Representation and implementation.}
We use all-MiniLM-L6-v2 to obtain 384-dimensional query and skill embeddings, mean-pooled to form the bundle representation. Each query--bundle pair is represented by a 776-dimensional feature vector consisting of the query embedding, bundle embedding, a four-dimensional bundle-condition indicator, and four GoS PPR summary statistics. The execution predictor is a $776\!\rightarrow\!256\!\rightarrow\!128\!\rightarrow\!64$ MLP with ReLU, dropout 0.3, and a logistic output head, trained using class-weighted binary cross-entropy and Adam ($10^{-3}$).

\paragraph{Baselines.}
We compare RADEG with \emph{Always-Execute}, which invokes the agent for every query--bundle pair; \emph{Random-Gate}, which assigns random execution scores;
\emph{PPR-Gate}, which ranks pairs using aggregate GoS relevance; and an \emph{Oracle} that executes exactly the positive-reward rollouts and serves only as an unattainable upper bound. 
Comparisons with a frozen execution head, a memory-corrected predictor, and a graph convolutional encoder are reported in Technical Supplement Section 1.3 (Complete Five-Split Results) and Section 1.4 (Additional Encoder Comparison).

\paragraph{Evaluation metrics.}
% Let $\mathcal{E}=\{(q_i,b_i,r_i)\}_{i=1}^{M}$
% denote a held-out evaluation set. For each query--bundle pair, RADEG
% produces an execution score
% \[
%     s_i=s_{\theta}(q_i,b_i)
% \]
% and, at threshold $\tau$, an execution decision
% \[
%     g_i(\tau)=\mathbb{I}[s_i\geq\tau].
% \]

We evaluate RADEG from both predictive and operational perspectives. We report the area under the receiver operating characteristic curve
(AUROC) and the area under the precision--recall curve (AUPRC). Both metrics are independent
of a particular decision threshold. AUROC measures how frequently a
positive-reward rollout is ranked above a zero-reward rollout, whereas
AUPRC emphasizes the ranking quality of the less frequent positive class.
Accuracy, precision, and recall at the default threshold $\tau=0.5$ are
reported as supplementary threshold-dependent metrics. Moreover, we define three metrics for the execution-gating setting. Let $M_{\mathrm{exec}}(\tau)
    =
    \sum_{i=1}^{M} g_i(\tau)$
denote the number of executed rollouts. We compute \emph{Calls Saved}, \emph{Reward Retained}, and \emph{Reward per Call}, respectively,
% \[
% \begin{aligned}
%     \emph{Calls Saved}(\tau)
%     &=
%     1-\frac{M_{\mathrm{exec}}(\tau)}{M},
%     \\[2mm]
%     \emph{Reward Retained}(\tau)
%     &=
%     \frac{\sum_{i=1}^{M}g_i(\tau)r_i}
%          {\sum_{i=1}^{M}r_i},
%     \\[2mm]
%     \emph{Reward per Call}(\tau)
%     &=
%     \frac{\sum_{i=1}^{M}g_i(\tau)r_i}
%          {M_{\mathrm{exec}}(\tau)}.
% \end{aligned}
% \]
\textbf{Calls Saved}$(\tau) = 1 - (M_{\mathrm{exec}}(\tau) / M)$ , 
\textbf{Reward Retained}$(\tau) = \sum_{i=1}^{M} g_i(\tau) r_i \big/ \sum_{i=1}^{M} r_i$ , 
and \textbf{Reward per Call}$(\tau) = \sum_{i=1}^{M} g_i(\tau) r_i \big/ M_{\mathrm{exec}}(\tau) $ .
Here, \emph{Calls Saved} is the fraction of agent executions avoided relative to Always-Execute; \emph{Reward Retained} is the fraction of the total
available verifier reward preserved by the gate; and \emph{Reward per Call} is the average verifier reward
obtained per executed rollout. The default $\tau=0.5$ is not tuned on the evaluation set; its robustness across a wide range of operating thresholds is verified in Technical Supplement Section 2.5 (Threshold Sensitivity).

\paragraph{Fixed-budget evaluation.}
A common threshold can lead different methods to execute different numbers
of calls. Following the cost--quality evaluation used in learned routing
\citep{ong2025routellm}, we therefore also compare methods under matched
execution budgets. For a budget $\beta\in(0,1]$, let $K_{\beta}=\lceil\beta M\rceil$
and let $\mathcal{I}_{\beta}$ contain the indices of the $K_{\beta}$
highest-scoring query--bundle pairs. We define $\emph{Reward retention}$ at budget
$\beta$ as
\[
    \emph{Reward retention}@\beta
    =
    \frac{\sum_{i\in\mathcal{I}_{\beta}}r_i}
         {\sum_{i=1}^{M}r_i},
\]
where $\emph{Reward retention}@\beta$ measures how much verifier reward a method
retains when it is allowed to execute only a fraction $\beta$ of the
Always-Execute calls.

Random-Gate results are averaged over 2,000 independent rankings. Differences in $\operatorname{RR}@\beta$ are assessed using 3,000 paired bootstrap resamples \citep{koehn2004statistical}. Bootstrap sampling is at the query level, preserving the four bundle variants of each query within the same resampled group.

\subsection{Experimental Results}

\paragraph{Comparison with Baselines on Unseen Queries}
\label{sec:main_results}
Table~\ref{tab:main_comparison} compares all methods under the same five leakage-free query-level splits. RADEG obtains the strongest threshold-free prediction performance among the deployable methods, with AUROC 0.717 and
AUPRC 0.570. Its improvement over PPR-Gate shows that upstream retrieval relevance alone does not reliably identify executions with non-zero verifier reward. RADEG also outperforms the Static and Memory-Corrected Heads,
demonstrating the value of the lightweight parametric online update. At $\tau=0.5$, RADEG skips 68\% of agent calls while retaining 61\% of the available verifier reward. This raises reward per executed call from 0.309
under Always-Execute to 0.483, a relative improvement of approximately 56\%. 
Since the methods operate at different execution rates under this threshold, 
the fixed-budget comparison below provides the cleaner cost-matched result. 
Complete classification metrics for all methods are reported in Technical Supplement Table 1 (Section 1.3). 

\paragraph{Robustness across query splits.}
To quantify sensitivity to the train--evaluation partition, we repeat the
complete RADEG procedure over 30 random query-level splits. RADEG obtains
AUROC $0.747\pm0.059$ and AUPRC $0.595\pm0.095$. At $\tau=0.5$, it skips
$68.1\%\pm6.9\%$ of calls while retaining $60.4\%\pm9.2\%$ of reward.
The five-split headline result therefore lies within the variability observed
under the larger uncertainty analysis. 
Full statistics and 95\% confidence intervals are reported in the Technical Supplement Section 2.1 (Thirty-Split Uncertainty Analysis).

\paragraph{Reward retention under fixed execution budgets.}
Table~\ref{tab:budget} compares ranking policies under matched budgets. RADEG retains more reward than both PPR-Gate and Random-Gate at every budget. At 20\% budget, it retains 40\% of total reward, versus 15\% for PPR-Gate and 19\% for Random-Gate. At 40\% budget, the corresponding values are 61\%, 40\%, and 41\%. The advantage is largest in the low-budget regime, where prioritization matters most.

\begin{table}[t]
\centering
\small
\setlength{\tabcolsep}{5pt}
\begin{tabular}{lcccc}
\toprule
& \multicolumn{4}{c}{Execution budget $\beta$} \\
\cmidrule(lr){2-5}
Method & 20\% & 40\% & 60\% & 80\% \\
\midrule
Oracle
& 65 & 100 & 100 & 100 \\
\textbf{RADEG}
& \textbf{40}
& \textbf{61}
& \textbf{73}
& \textbf{87} \\
PPR-Gate
& 15 & 40 & 57 & 81 \\
Random-Gate
& 19 & 41 & 61 & 77 \\
\bottomrule
\end{tabular}
\caption{
Reward retained, $\operatorname{RR}@\beta$ (\%), under matched execution budgets in the common five-split comparison. Always-Execute is the normalization reference; Oracle is an unattainable post-hoc upper bound.
}
\label{tab:budget}
\end{table}

As a complementary robustness analysis, we perform 3,000 paired bootstrap resamples at the query level, keeping all bundle variants of a query in the same resampled group \citep{koehn2004statistical}. RADEG's advantage over
both baselines remains positive at every evaluated budget ($p<0.001$). Effect sizes and confidence intervals are provided in the Technical Supplement Section 2.2 (Fixed-Budget Statistical Analysis).

% \begin{figure}[t]
%     \centering
%     \includegraphics[width=\columnwidth]
%     {figures/budget_curve_matched_protocol.pdf}
%     \caption{
%     Reward retained under matched execution budgets. RADEG provides its
%     largest advantage in the low-budget regime.
%     }
%     \label{fig:budget}
% \end{figure}
\paragraph{Cross-agent adaptation.}
We examine whether the representation learned from Sonnet executions can support gating for different downstream execution configurations. The gate does not receive the target-agent identity. For each target, the encoder is
frozen and the execution head is initialized from the same Sonnet-trained checkpoint. The target stream is then evaluated prequentially: each pair is scored before its target reward is revealed, and only previously observed
target rewards may affect subsequent predictions. The head is reset between targets, and results are averaged over five runs. 

\begin{table}[t]
\centering
\small
\setlength{\tabcolsep}{4pt}
\resizebox{\columnwidth}{!}{
\begin{tabular}{lccccc}
\toprule
Target agent
& $n$
& Pos.
& AUROC
& Calls saved
& Reward retained \\
\midrule
Qwen3.7-Max
& 50 & 33
& $0.733\pm0.024$
& 34\% & 73\% \\
DeepSeek-Chat
& 62 & 25
& $0.628\pm0.018$
& 58\% & 53\% \\
GLM-5.2
& 41 & 30
& $0.752\pm0.029$
& 27\% & 83\% \\
\bottomrule
\end{tabular}
}
\caption{
Per-target cross-agent adaptation from the Sonnet-trained checkpoint.
}
\label{tab:cross_agent}
\end{table}
Table~\ref{tab:cross_agent} reports the primary cross-agent adaptation results.
RADEG obtains AUROCs of $0.733\pm0.024$ on Qwen3.7-Max,
$0.628\pm0.018$ on DeepSeek-Chat, and $0.752\pm0.029$ on GLM-5.2. These results provide evidence that the source-trained representation can
support utility prediction across multiple downstream execution configurations, although performance is target-dependent. 
Complete cross-agent transfer results under static transfer, per-target adaptation, and continual adaptation, along with protocol details and limitations, are reported in Technical Supplement Section 3.1-3.3.

\paragraph{Generalization across retrievers and bundle sizes.}
We further evaluate whether RADEG depends on a particular upstream retriever or bundle size. Using DeepSeek-V4-Pro as the execution model, we construct query--bundle pairs with four retrieval strategies---hybrid retrieval, BM25, embedding similarity, and a gold-distractor construction---and evaluate both the original top-8 bundles and top-7 variants obtained by removing one skill.
The gate is trained and evaluated under the same 80/20 query-level split protocol, with results averaged over five runs.

As shown in Table \ref{tab:retriever_bundle_generalization}, RADEG remains effective across the three standard retrievers. On top-8 bundles, AUROC is $0.750\pm0.032$ for hybrid retrieval and $0.683\pm0.041$ for both BM25 and embedding retrieval. Reducing the bundle
size from eight to seven skills results in only a small overall change, from $0.672\pm0.026$ to $0.645\pm0.005$, suggesting that the learned utility
signal is not tied to the exact bundle cardinality. Performance is consistently stronger for hybrid, BM25, and embedding retrieval than for the gold-distractor construction, whose combined AUROC is $0.542\pm0.013$. This indicates that RADEG generalizes
across naturally produced retrieval bundles, while deliberately constructed
distractor bundles induce a more substantial distribution shift.
\begin{table}[t]
\centering
\scriptsize
\setlength{\tabcolsep}{5pt}
\begin{tabular}{lccc}
\toprule
Retriever & Top-8 & Top-7 & Combined \\
\midrule
Hybrid & $0.750\pm0.032$ & $0.723\pm0.005$ & $\mathbf{0.734\pm0.003}$ \\
BM25 & $0.683\pm0.041$ & $0.645\pm0.011$ & $0.654\pm0.012$ \\
Embedding & $0.683\pm0.057$ & $0.662\pm0.006$ & $0.676\pm0.008$ \\
Distractor & $0.300\pm0.053$ & $0.498\pm0.003$ & $0.542\pm0.013$ \\
\midrule
Overall & $0.672\pm0.026$ & $0.645\pm0.005$ & $0.662\pm0.005$ \\
\bottomrule
\end{tabular}
\caption{
RADEG utility prediction across retrievers and bundle sizes.
Results are AUROC mean$\pm$standard deviation over five runs.
}
\label{tab:retriever_bundle_generalization}
\end{table}
\paragraph{Learning under selective feedback.}
Full-information replay is more informative than deployment because it reveals labels even for simulated skips. We therefore evaluate $\varepsilon$-greedy exploration under selective feedback. When RADEG would skip a pair, the system executes it with probability $\varepsilon$ and
observes its reward. Thus, $\varepsilon=0$ corresponds to pure selective feedback, whereas $\varepsilon=1$ recovers full label availability. Additional results under selective feedback and different exploration rates are reported in the Technical Supplement Section 2.3 (Selective Feedback and Exploration). Performance improves as more skipped calls are explored, suggesting that missing labels are
a cause of degradation under selective feedback. Exploration
partially corrects this bias by revealing rewards for some calls; however, it also increases the number of agent executions.

\paragraph{Feature ablation.}
We remove each feature block and repeat the evaluation over 30 query-level splits. As shown in Table \ref{tab:feature_ablation}, removing the bundle embedding produces the largest numerical decrease in
mean AUROC, from 0.744 to 0.730, suggesting that bundle semantics provides a useful utility signal. Removing the PPR statistics has almost no effect, and removing the explicitly encoded bundle condition slightly increases the mean. All differences are small relative to split-level variability, so we do not claim a statistically significant ordering among feature groups. The results
reduce the likelihood that RADEG merely memorizes the perturbation category or reproduces the upstream relevance score. Representative successful and failure cases, including false positive and false negative gating decisions, are analyzed in the Technical Supplement section 4.2 (Case Studies and Error Analysis).

\begin{table}[t!]
\centering
\small
\begin{tabular}{lc}
\toprule
Input representation
& AUROC $\uparrow$ \\
\midrule
Full input
& $0.744\pm0.070$ \\
w/o query embedding
& $0.739\pm0.067$ \\
w/o bundle embedding
& $0.730\pm0.058$ \\
w/o bundle-condition feature
& $0.751\pm0.067$ \\
w/o PPR statistics
& $0.742\pm0.068$ \\
\bottomrule
\end{tabular}
\caption{
Feature ablation over 30 query-level splits.
Results are mean$\pm$standard deviation.
}
\label{tab:feature_ablation}
\end{table}

\section{Conclusion}

This paper separates two decisions often conflated in skill-based agents:
retrieving a relevant bundle and deciding whether executing it is worthwhile.
Controlled perturbations show that GoS-selected bundles are reward-sensitive,
while aggregate PPR features provide little predictive signal for non-zero
reward. We therefore introduced RADEG, a post-retrieval execution gate trained
from query--bundle--reward rollouts. Under held-out evaluation, RADEG achieves an AUROC of 0.717, retaining 61\% of reward while skipping 68\% of calls at the default threshold, and outperforming PPR and random ranking under matched budgets.

These experimental results support execution-utility prediction as a lightweight, retrieval-agnostic complement to skill retrieval, with applicability beyond GoS to different relevance-based retrieval pipelines. RADEG leaves both the retriever and the downstream agent unchanged, making it applicable to different skill-selection pipelines with low integration cost. 
The current study remains limited by the scale of collected rollouts, reliance on logged verifier feedback, the scope of controlled perturbation types, and incomplete feedback when skipped executions are not explored.
Future work therefore will evaluate RADEG on larger and more diverse skill libraries, develop more effective learning strategies under selective feedback, and measure end-to-end monetary and latency savings in real deployment settings.

\medskip

\bibliography{aaai2027}

\appendix
\section*{Technical Supplement}
This Technical Supplement provides additional experimental details, analyses,
and robustness studies supporting the main paper
"From Relevance to Execution Utility: Reward-Aware Dynamic Execution Gating
for Skill-Based LLM Agents."

Technical Supplement (Complete Evaluation Details) presents the complete
evaluation protocols, including the experimental environment, compared
methods, full classification and operational metrics, and additional encoder
comparisons that complement the compact experimental results reported in the
main paper.

Technical Supplement (Robustness and Operating-Point Analysis) provides
additional robustness analyses, including uncertainty evaluation over multiple
query-level splits, bootstrap significance tests under fixed execution
budgets, selective-feedback evaluation, exploration studies, permutation
testing for PPR relevance, and execution-threshold sensitivity analyses.

Technical Supplement (Extended Cross-Agent Evaluation) reports additional
experiments across different downstream agent configurations, including
per-target evaluation results, continual adaptation settings, and discussions
of protocol limitations. These experiments further examine whether RADEG's
execution-utility prediction remains effective when the underlying execution
agent changes.

Technical Supplement (Additional Dataset Analysis) presents additional
statistics of the collected execution dataset and analyzes the effects of
controlled bundle perturbations. Section 4.2 further provides representative
case studies and error analyses that characterize execution patterns and false
gating decisions.

Technical Supplement (Runtime Analysis) reports end-to-end execution time
across all evaluated agent backbones and measures the computational overhead
of RADEG itself, demonstrating that the execution gate introduces negligible
latency relative to downstream agent execution while remaining practical for
online deployment.

Technical Supplement (Extended Cross-Benchmark Generalization) evaluates RADEG
on three independent tool-agent benchmarks beyond the original SkillBench
setting. This section studies whether execution-utility prediction generalizes
across different tool ecosystems and execution environments under controlled
tool-bundle constructions.

These additional experiments and analyses provide further evidence
that RADEG predicts execution utility beyond retrieval relevance, while
demonstrating its robustness, generalization ability, and practical
applicability across different evaluation settings.
\section{Complete Evaluation Details}
\subsection{Experimental Environment}

Unless otherwise specified, all offline training, execution-gating inference,
and data analysis were conducted on a local workstation equipped with an
Apple M2 Pro processor and 16\,GB unified memory running macOS 15.6.1.
The lightweight execution gate was evaluated on CPU only. The implementation
uses Python 3.13.7 and PyTorch 2.10.0. Query and skill embeddings are
generated using \texttt{all-MiniLM-L6-v2} and cached before online execution.
Agent rollouts are executed in Docker containers using the corresponding
execution environments of each evaluated agent.
\subsection{Compared Methods and Evaluation Protocol}

The common comparison evaluates all methods under the same five
leakage-free query-level splits. All available bundle variants associated
with a query are assigned to the same partition. Unless otherwise stated,
threshold-dependent operating metrics use $\tau=0.5$.

We compare RADEG with the following execution policies.
\textbf{Always-Execute} executes every retrieved bundle and provides the
reward-normalization reference.
\textbf{Random-Gate} assigns an uninformed random ranking.
\textbf{PPR-Gate} uses aggregate PPR relevance as its execution score,
testing whether the upstream retriever's relevance signal can also serve as
an execution-utility policy.
\textbf{Static Head} uses the offline-trained utility encoder and logistic
head without online updates.
\textbf{Memory-Corrected Head} combines the static prediction with the
positive-label frequency among the $k=5$ cosine-nearest stored
representations:
\[
    s
    =
    \alpha s_{\mathrm{static}}
    +(1-\alpha)s_{\mathrm{mem}},
    \qquad
    \alpha=0.7.
\]
For classification, \textbf{Oracle} uses the ground-truth execution label as
its score. For fixed-budget ranking, it orders calls using their true
verifier rewards. Oracle is a post-hoc upper bound and is not deployable.

\subsection{Complete Five-Split Results}

Table~\ref{tab:full_five_split} reports the complete classification and
operational metrics underlying the compact comparison in the main paper.

\begin{table*}[t]
\centering
\small
\resizebox{\textwidth}{!}{%
\begin{tabular}{lcccccccc}
\toprule
Method
& Acc.
& AUROC
& AUPRC
& Prec.
& Recall
& Calls saved
& Reward retained
& Reward/call \\
\midrule
Always-Execute
& -- & -- & -- & -- & --
& 0\% & 100\% & 0.309 \\

Random-Gate
& 0.564 & 0.500 & 0.345 & 0.346 & 0.297
& 71\% & 30\% & 0.313 \\

PPR-Gate
& 0.447 & 0.447 & 0.335 & 0.305 & 0.482
& 46\% & 51\% & 0.292 \\

Static Head
& 0.567 & 0.517 & 0.375 & 0.350 & 0.285
& 73\% & 27\% & 0.308 \\

Memory-Corrected Head
& 0.574 & 0.608 & 0.440 & 0.363 & 0.291
& 73\% & 28\% & 0.321 \\

\textbf{RADEG}
& \textbf{0.706}
& \textbf{0.717}
& \textbf{0.570}
& \textbf{0.569}
& \textbf{0.560}
& 68\%
& \textbf{61\%}
& \textbf{0.483} \\

Oracle
& 1.000 & 1.000 & 1.000 & 1.000 & 1.000
& 65\% & 100\% & 0.892 \\
\bottomrule
\end{tabular}%
}
\caption{
Complete comparison under the common five-split query-level protocol.
Operational metrics use $\tau=0.5$.
}
\label{tab:full_five_split}
\end{table*}

\subsection{Additional Encoder Comparison}
\label{app:gcn}
Table~\ref{tab:gcn} summarizes the comparison between RADEG and three representative alternatives under the common five-split evaluation protocol.
Because GoS exposes a skill dependency graph, we additionally replace the
MLP utility encoder with a two-layer graph convolutional network.
The evaluated skill subgraph contains 200 nodes and 126 edges. The GCN
propagates 384-dimensional node features through
$384\!\rightarrow\!128\!\rightarrow\!64$ layers before combining them with
the query--bundle representation.

\begin{table}[t]
\centering
\small
\begin{tabular}{lcc}
\toprule
Configuration
& AUROC $\uparrow$
& Reward retained $\uparrow$ \\
\midrule
GCN Gate
& 0.437 & 18\% \\

Static Head
& 0.517 & 27\% \\

Memory-Corrected Head
& 0.608 & 28\% \\

\textbf{RADEG}
& \textbf{0.717}
& \textbf{61\%} \\
\bottomrule
\end{tabular}
\caption{
Encoder comparison under the common five-split protocol.
}
\label{tab:gcn}
\end{table}

The tested GCN performs below the MLP-based configurations. One plausible
explanation is that the graph is sparse relative to the number of available
execution labels, leaving insufficient supervision for message passing to
exploit the graph structure reliably. This result applies only to the tested
GCN construction and does not imply that graph information is generally
unhelpful.

%%%%%%%%%%%%%%%%%%%%%%%%%%%%%%%%%%%%%%%%%%%%%%%%%%%%%%%%%%%%
\section{Robustness and Operating-Point Analysis}
\label{app:robustness}

\subsection{Thirty-Split Uncertainty Analysis}
\label{app:uncertainty}
Table~\ref{tab:uncertainty} summarizes RADEG over 30 independently generated
query-level splits. Each split satisfies the same leakage-free protocol used
throughout the paper, and operational metrics are computed using the default
decision threshold $\tau=0.5$.
The common baseline comparison uses five shared query-level splits so that
all methods are evaluated under the same protocol. We separately repeat the
complete RADEG training and evaluation procedure over 30 random query-level
splits to assess the sensitivity of the proposed method to the
train--evaluation partition. This analysis evaluates RADEG's stability and
does not repeat the full baseline comparison.

\begin{table}[t]
\centering
\small
\begin{tabular}{lccc}
\toprule
Metric
& Mean
& Std.
& 95\% CI \\
\midrule
AUROC
& 0.747
& 0.059
& $[0.726,0.767]$ \\

AUPRC
& 0.595
& 0.095
& $[0.561,0.628]$ \\

Calls saved
& 0.681
& 0.069
& $[0.657,0.705]$ \\

Reward retained
& 0.604
& 0.092
& $[0.570,0.635]$ \\

Reward per call
& 0.521
& 0.079
& $[0.493,0.549]$ \\
\bottomrule
\end{tabular}
\caption{
RADEG performance over 30 leakage-free query-level splits.
Operational metrics use $\tau=0.5$.
}
\label{tab:uncertainty}
\end{table}

The five-split AUROC of 0.717 lies within the variability observed over the
larger collection of query partitions. The 30-split results therefore
indicate that RADEG's predictive and operational performance is not driven by
one favorable split.

\subsection{Fixed-Budget Statistical Analysis}
\label{app:budget_bootstrap}

For the random-ranking reference, Random-Gate results are averaged over
2,000 independent random rankings. We further perform 3,000 paired bootstrap
resamples of the pooled out-of-fold predictions. Resampling is conducted at
the query level, keeping all available bundle variants of a query within the
same resampled group.

Table~\ref{tab:budget_significance} reports the resulting differences in
reward retention. Because this bootstrap analysis uses pooled out-of-fold
predictions, its mean differences need not equal the direct arithmetic
difference between the rounded five-split means reported in the main table.

\begin{table}[t]
\centering
\scriptsize
\begin{tabular}{lcc}
\toprule
Budget
& $\Delta$ vs.\ Random-Gate
& $\Delta$ vs.\ PPR-Gate \\
\midrule
20\%
& $20.4\ [17.1,23.9]$
& $26.7\ [22.9,30.4]$ \\

40\%
& $27.9\ [23.7,32.1]$
& $31.8\ [27.5,36.1]$ \\

60\%
& $18.8\ [14.9,22.6]$
& $25.2\ [21.1,29.5]$ \\

80\%
& $9.2\ [6.2,12.2]$
& $8.9\ [5.5,12.2]$ \\
\bottomrule
\end{tabular}
\caption{
Paired improvement in reward retention. Values are percentage-point
differences with bootstrap 95\% confidence intervals. All one-sided tests
have $p<0.001$.
}
\label{tab:budget_significance}
\end{table}

\subsection{Selective Feedback and Exploration}
\label{app:exploration}
Table~\ref{tab:selective_main} reports the effect of different
$\varepsilon$-greedy exploration rates on online utility prediction.
Under selective feedback, rewards are observed only for calls that are
actually executed. We simulate explicit exploration by overriding an
otherwise skipped decision with probability $\varepsilon$ and revealing its
verifier reward.
\begin{table}[t]
\centering
\small
\begin{tabular}{lc}
\toprule
Exploration rate $\varepsilon$
& AUROC $\uparrow$ \\
\midrule
0.00 & $0.557\pm0.077$ \\
0.20 & $0.608\pm0.086$ \\
0.50 & $0.674\pm0.095$ \\
1.00 & $0.745\pm0.065$ \\
\bottomrule
\end{tabular}
\caption{
Effect of $\varepsilon$-greedy exploration on online utility prediction.
Results are mean$\pm$standard deviation.
}
\label{tab:selective_main}
\end{table}

\begin{table}[t]
\centering
\small
\begin{tabular}{lc}
\toprule
Exploration rate $\varepsilon$
& AUROC $\uparrow$ \\
\midrule
0.00
& $0.557\pm0.077$ \\

0.05
& $0.582\pm0.101$ \\

0.10
& $0.591\pm0.091$ \\

0.20
& $0.608\pm0.086$ \\

0.50
& $0.674\pm0.095$ \\

1.00
& $0.745\pm0.065$ \\
\bottomrule
\end{tabular}
\caption{
Prequential AUROC under selective feedback and
$\varepsilon$-greedy exploration. Results are reported as
mean$\pm$standard deviation.
}
\label{tab:exploration_full}
\end{table}
To further characterize the effect of exploration, we additionally evaluate
a denser sweep of exploration rates ranging from $\varepsilon=0$ to
$\varepsilon=1$. Table~\ref{tab:exploration_full} reports the complete
results.

The increasing mean AUROC is consistent with missing labels being an
important source of degradation under selective feedback. We do not claim
that adjacent exploration rates differ significantly, as the split-level
variability remains substantial. Moreover, the table isolates predictive
performance and does not account for the additional agent calls consumed by
exploration.

\subsection{PPR permutation test}
\label{app:ppr_test}

We shuffle reward labels 10,000 times and recompute the AUROC of aggregate
PPR relevance. The observed AUROC is 0.435, while the permutation
distribution has mean 0.499 and 95\% interval $[0.427,0.572]$.
The observed value lies inside the null interval, providing no evidence that
aggregate PPR relevance ranks reward-positive executions more effectively
than an uninformative ordering in this dataset.

The value differs slightly from the five-split PPR-Gate AUROC in
Table~\ref{tab:full_five_split} because the permutation test is computed once
over the pooled source rollouts, whereas the table reports an average over
held-out splits.
\subsection{Threshold Sensitivity}
\label{app:threshold_sensitity}
Table~\ref{tab:threshold_curve} reports pooled out-of-fold operating points
under different execution thresholds. Since RADEG outputs the probability of
positive execution utility, the threshold $\tau$ controls the trade-off between
execution savings and retained reward.

\begin{table*}[t]
\centering
\small
\begin{tabular}{lcccccc}
\toprule
$\tau$
& Calls saved
& Reward retained
& Precision
& Recall
& F1
& Accuracy \\
\midrule
0.1
& 64\%
& 65\%
& 0.587
& 0.669
& 0.626
& 0.747 \\

0.2
& 65\%
& 64\%
& 0.601
& 0.657
& 0.628
& 0.754 \\

0.3
& 67\%
& 63\%
& 0.608
& 0.644
& 0.626
& 0.756 \\

0.4
& 67\%
& 62\%
& 0.613
& 0.634
& 0.623
& 0.758 \\

\textbf{0.5} & \textbf{68\%} & \textbf{61\%} & \textbf{0.616} & \textbf{0.629} & \textbf{0.622} & \textbf{0.759} \\

0.6
& 68\%
& 61\%
& 0.622
& 0.623
& 0.622
& 0.761 \\

0.7
& 69\%
& 60\%
& 0.625
& 0.614
& 0.619
& 0.761 \\

0.8
& 70\%
& 59\%
& 0.625
& 0.599
& 0.612
& 0.760 \\

0.9
& 72\%
& 55\%
& 0.627
& 0.565
& 0.595
& 0.756 \\
\bottomrule
\end{tabular}
\caption{
RADEG operating points under different execution thresholds.
}
\label{tab:threshold_curve}
\end{table*}

The results show that RADEG is robust across a wide range of thresholds.
Increasing $\tau$ gradually improves execution savings and precision while
reducing retained reward and recall, following the expected precision--recall
trade-off. The default threshold $\tau=0.5$ provides a balanced operating point:
it achieves near-best accuracy (0.759), competitive F1 (0.622), retains 61\% of
available reward, and skips 68\% of executions. Therefore, $\tau=0.5$ is used as a fixed operating point chosen for its balanced trade-off between execution savings and retained reward, rather than optimized for the evaluation set.
% \paragraph{Threshold sensitivity.}
% Table~\ref{tab:threshold_curve} reports pooled out-of-fold operating points
% for different execution thresholds.

% \begin{table}[t]
% \centering
% \small
% \caption{
% RADEG operating points under different execution thresholds.
% }
% \label{tab:threshold_curve}
% \begin{tabular}{lcc}
% \toprule
% Threshold
% & Calls executed
% & Reward retained \\
% \midrule
% 0.05
% & 39\%
% & 68\% \\

% 0.20
% & 34\%
% & 63\% \\

% 0.30
% & 33\%
% & 61\% \\

% 0.40
% & 32\%
% & 59\% \\

% 0.50
% & 31\%
% & 58\% \\

% 0.60
% & 31\%
% & 57\% \\

% 0.70
% & 30\%
% & 56\% \\

% 0.80
% & 29\%
% & 55\% \\
% \bottomrule
% \end{tabular}
% \end{table}

% Lower thresholds execute more calls and retain more reward, whereas higher
% thresholds provide greater execution savings. The operating threshold should
% therefore be selected according to an explicit deployment utility function
% rather than treated as a universal constant.

%%%%%%%%%%%%%%%%%%%%%%%%%%%%%%%%%%%%%%%%%%%%%%%%%%%%%%%%%%%%
\section{Extended Cross-Agent Evaluation}
\label{app:cross_agent}

\subsection{Protocol and Target Composition}

The gate consists of a utility encoder $\phi$ and logistic execution head
$h$. We consider three cross-agent protocols.

In \emph{static transfer}, both $\phi$ and $h$ remain frozen throughout the
target evaluation. In \emph{per-target adaptation}, used for the primary
cross-agent results, $\phi$ remains frozen while $h$ is reset to the same
Sonnet-trained checkpoint for each target agent and then updated
prequentially using that target's verifier feedback. In
\emph{continual adaptation}, a single head is preserved across a fixed
sequence of target-agent streams without resetting.

Static and prequential AUROCs answer different questions. Static AUROC
evaluates one fixed scoring function, whereas online AUROC aggregates
predictions made by an evolving head. Every prediction is produced before its
own label is revealed, but the static and online values should not be
interpreted as the beginning and end of one learning curve.

We treat a target as sufficiently supported for the primary AUROC analysis
when it contains at least ten positive and ten zero-reward examples.

\subsection{Complete Cross-Agent Results}

Table~\ref{tab:all_cross_agent} consolidates the static, per-target, and
continual cross-agent results. Static results are reported only for targets
for which protocol-matched frozen evaluations are available.

\begin{table*}[t]
\centering
\scriptsize
\resizebox{\textwidth}{!}{%
\begin{tabular}{lcccccc}
\toprule
Target agent
& $n$
& Positive
& Frozen
& Per-target
& Continual
& Status \\
\midrule
\textbf{Qwen3.7-Max}
& 50
& 33
& 0.807
& $\mathbf{0.733\pm0.024}$
& $0.721\pm0.029$
& Primary \\

\textbf{DeepSeek-Chat}
& 62
& 25
& 0.733
& $\mathbf{0.628\pm0.018}$
& $0.641\pm0.025$
& Primary \\

\textbf{GLM-5.2}
& 41
& 30
& --
& $\mathbf{0.752\pm0.029}$
& $0.757\pm0.032$
& Primary \\

Claude Sonnet-4
& 27
& 3
& --
& $0.822\pm0.024$
& $0.869\pm0.021$
& Underpowered \\

GPT-4o
& 20
& 2
& --
& $0.606\pm0.099$
& $0.606\pm0.099$
& Underpowered \\

Gemini 2.5 Pro
& 30
& 0
& --
& N/A
& N/A
& Single class \\

Gemini 2.5 Flash
& 21
& 0
& --
& N/A
& N/A
& Single class \\

GLM-4.6
& 17
& 2
& --
& $0.533\pm0.059$
& $0.587\pm0.072$
& Underpowered \\

Qwen3-Max
& 2
& 1
& --
& $1.000\pm0.000$
& $1.000\pm0.000$
& Underpowered \\

Kimi-K2
& 2
& 2
& --
& N/A
& N/A
& Single class \\
\bottomrule
\end{tabular}%
}
\caption{
Complete cross-agent results. Frozen evaluation keeps the full source gate
fixed; per-target adaptation resets the head for each target; continual
adaptation preserves one head across target-agent streams. Frozen and online
AUROCs correspond to different protocols and should not be directly
subtracted.
}
\label{tab:all_cross_agent}
\end{table*}

The underpowered and single-class rows are reported for completeness but are
not used to support the primary cross-agent conclusion. In particular, high
AUROCs obtained from only one to three positive examples should not be
interpreted as reliable transfer estimates.

We additionally evaluate the subset of pairs for which the source Sonnet
execution receives non-zero reward. The corresponding per-target AUROCs are
$0.733\pm0.024$ for Qwen3.7-Max,
$0.660\pm0.028$ for DeepSeek-Chat, and
$0.752\pm0.029$ for GLM-5.2. Because this subset is selected using
source-agent outcomes, it measures ranking among calls that were already
useful to the source agent rather than performance on the complete target
distribution.

\subsection{Cross-Agent Limitations}

The target streams contain multiple bundle variants associated with the same
query. Although each rollout is scored before its own label is revealed,
feedback from an earlier bundle variant may influence predictions for later
variants of the same query. The protocol therefore avoids direct
test-label leakage but does not represent a strict
single-bundle-per-query deployment. A stricter evaluation should score all
variants of a query before revealing any corresponding labels, or evaluate
only one retrieved bundle per query.

The source rollouts use a Claude-Code-style execution harness, whereas the
non-Claude target agents use a lighter-weight agent scaffold. Differences in
target rewards may therefore reflect both execution-model behavior and
scaffold behavior. We consequently interpret the results as adaptation
across complete downstream execution configurations rather than as a
controlled comparison of model weights alone.

%%%%%%%%%%%%%%%%%%%%%%%%%%%%%%%%%%%%%%%%%%%%%%%%%%%%%%%%%%%%
\section{Additional Dataset Analysis}
\label{app:data_analysis}

\subsection{Dataset Composition and Perturbation Effects}

The source gate dataset contains 72 complete four-condition query groups,
corresponding to 288 rollouts. Query-level splitting keeps every available rollout associated
with a query in the same partition.

Of the 288 rollouts, 194 receive zero reward, 24 receive partial reward
($0<r<1$), and 70 receive full reward ($r=1$). Partial rewards constitute
25.5\% of all positive examples, supporting the use of
\[
    y=\mathbb{I}[r>0]
\]
as the binary execution-utility label rather than restricting positive
utility to full-reward executions.

Table~\ref{tab:perturbation_distribution} reports the task-level effects of
the three controlled bundle perturbations relative to the original GoS
bundle.

\begin{table}[t]
\centering
\scriptsize
\resizebox{\columnwidth}{!}{%
\begin{tabular}{lcccccc}
\toprule
Perturbation
& Up
& Down
& Same
& Changed
& Mean $\Delta$
& $0\!\to\!+\,/\,+\!\to\!0$ \\
\midrule
Delete Top
& 8
& 7
& 57
& 21\%
& $+0.019$
& 7 / 5 \\

Add Irrelevant
& 6
& 6
& 60
& 17\%
& $-0.016$
& 4 / 5 \\

Replace Similar
& 8
& 6
& 58
& 19\%
& $+0.010$
& 6 / 5 \\
\bottomrule
\end{tabular}%
}
\caption{
Task-level effects of each perturbation relative to GoS Original.
}
\label{tab:perturbation_distribution}
\end{table}

No perturbation is uniformly beneficial. These interventions expose the
sensitivity of execution outcomes to bundle composition and are not proposed
as fixed replacement policies.

\subsection{Case Studies and Error Analysis}

Table~\ref{tab:case_studies} illustrates three representative execution
patterns.

\begin{table}[t]
\centering
\scriptsize
\begin{tabular}{p{3.0cm}cccc}
\toprule
Query
& Orig.
& Del.
& Add
& Repl. \\
\midrule
adaptive-cruise-control
& 0
& 0
& 0
& 1 \\

drone-planning-control
& 0.3
& 0
& 0.5
& 0.4 \\

ada-bathroom-plan-repair
& 0
& 0
& 0
& 0 \\
\bottomrule
\end{tabular}
\caption{
Representative verifier rewards across bundle conditions.
}
\label{tab:case_studies}
\end{table}

For \texttt{adaptive-cruise-control}, the original GoS execution fails while
the similar-replacement bundle succeeds, illustrating a zero-to-positive
transition. For \texttt{drone-planning-control}, different perturbations
either reduce or improve an already partially successful execution,
demonstrating that bundle changes are not uniformly beneficial. For
\texttt{ada-bathroom-plan-repair}, every tested bundle fails, making the
query a natural candidate for execution skipping.

Across the repeated five-split evaluation, pooled decisions contain 60 false
positives and 60 false negatives. These counts aggregate predictions from
repeated splits and must not be interpreted as 120 distinct rollouts.
Representative missed positive executions include
\texttt{mars-clouds-clustering},
\texttt{earthquake-phase-association},
\texttt{data-to-d3}, and
\texttt{gravitational-wave-detection}.
The current sample is insufficient for a reliable domain-level error
taxonomy. A larger evaluation should distinguish representation error,
limited historical coverage, stochastic execution variation, and
verifier-specific failure modes.
\section{Runtime Analysis}
\label{app:runtime}

Because RADEG is designed to reduce unnecessary agent executions, we
additionally evaluate both the end-to-end runtime of downstream agent
execution and the computational overhead introduced by the execution gate.

Table~\ref{tab:agent_runtime} reports the average end-to-end runtime,
measured from agent invocation to verifier completion, for all evaluated
execution agents. Runtime varies substantially across agents, ranging from
60\,s for Kimi-K2 to over 1,500\,s for GLM-5.2.

\begin{table}[t]
\centering
\small
\begin{tabular}{lc}
\toprule
Execution agent & Runtime (s) \\
\midrule
Qwen3.7-Max & 1431 \\
DeepSeek-Chat & 521 \\
GLM-5.2 & 1542 \\
Claude Sonnet-4 & 560 \\
GPT-4o & 274 \\
Gemini 2.5 Pro & 506 \\
Gemini 2.5 Flash & 514 \\
GLM-4.6 & 485 \\
Qwen3-Max & 396 \\
Kimi-K2 & 60 \\
\bottomrule
\end{tabular}
\caption{Average end-to-end runtime of each execution agent, measured from
agent invocation to verifier completion.}
\label{tab:agent_runtime}
\end{table}

We further measure the computational overhead of RADEG itself on a
single-thread CPU using 5,000 repeated measurements (median over three
runs). Query and skill embeddings are pre-computed and cached; therefore,
the reported runtime corresponds only to online execution gating.
Table~\ref{tab:gate_runtime} summarizes the runtime of each component of the
execution gate.
\begin{table}[t]
\centering
\small
\begin{tabular}{lc}
\toprule
Operation & Runtime \\
\midrule
Feature construction & 22 $\mu$s \\
MLP forward pass & 50 $\mu$s \\
Single prediction & 75 $\mu$s \\
Prediction + online update & 131 $\mu$s \\
Batched prediction (per sample) & 2.0 $\mu$s \\
\bottomrule
\end{tabular}
\caption{Runtime of the RADEG execution gate.}
\label{tab:gate_runtime}
\end{table}

The execution gate introduces negligible computational overhead compared
with downstream agent execution. A complete execution-gating decision
requires approximately $75\,\mu$s ($0.075$\,ms), whereas the fastest
evaluated execution agent requires approximately $60$\,s and the primary
cross-agent evaluation on DeepSeek-Chat requires approximately $521$\,s.
Consequently, the runtime of RADEG is over six orders of magnitude smaller
than the execution it decides whether to invoke. Even processing all 288
rollouts in the source dataset requires only approximately 22\,ms, showing
that the execution gate is lightweight enough to be deployed online without
becoming a practical bottleneck.
%%%%%%%%%%%%%%%%%%%%%%%%%%%%%%%%%%%%%%%%%%%%%%%%%%%%%%%%%%%%
\section{Extended Cross-Benchmark Generalization}
\label{app:cross_benchmark}

To further evaluate whether RADEG depends on a specific skill retrieval
pipeline, we conduct additional experiments on three independent tool-agent
benchmarks: $\tau$-Bench, AgentDojo, and ToolSandbox. These benchmarks contain
different tool ecosystems, task distributions, and execution environments.
This analysis examines whether RADEG's execution-utility prediction capability
generalizes beyond the original SkillBench setting.

\subsection{Benchmark Construction and Evaluation Protocol}

Unlike the original SkillBench evaluation, where skill bundles are obtained
through Graph-of-Skills retrieval, these benchmarks provide native tool
interfaces. For each benchmark, we construct query--tool bundles under a
controlled setting. Specifically, we create three bundle conditions:
\textbf{Gold}, containing task-relevant tools; \textbf{Halfmix}, containing a
mixture of relevant and irrelevant tools; and \textbf{Distractor}, containing
irrelevant tools only.

All three conditions contain the same number of tools. Therefore, the
different execution outcomes are caused by tool composition rather than the
number of available tools. This design eliminates bundle-size as a potential
confounding factor and allows us to evaluate whether RADEG captures
query--tool compatibility.

The execution rewards are collected using DeepSeek-V4-Pro through the
official API endpoint. For each benchmark, we collect 60 executions, including
20 queries under each bundle condition, resulting in 180 additional
query--tool execution records. The input representation follows the original
RADEG design. Query and tool descriptions are encoded using
\texttt{all-MiniLM-L6-v2}. Tool representations are mean-pooled and combined
with the query representation. Since these benchmarks do not provide
Graph-of-Skills metadata, graph-related features are set to zero.

\subsection{Execution Outcome Analysis}

We first analyze whether tool composition affects downstream execution
utility. Table~\ref{tab:cross_benchmark_collection} reports execution success
under the three controlled bundle conditions.

\begin{table}[t]
\centering
\small
\setlength{\tabcolsep}{1pt}
\begin{tabular}{lcccc}
\toprule
Benchmark
& Gold
& Halfmix
& Distractor
& $n$ \\
\midrule
$\tau$-Bench (Retail)
& 13/20
& 2/20
& 2/20
& 60 \\

AgentDojo (Bank/Travel/Workspace)
& 17/20
& 8/20
& 3/20
& 60 \\

ToolSandbox (Apple)
& 20/20
& 6/20
& 6/20
& 60 \\
\bottomrule
\end{tabular}
\caption{
Execution success under controlled tool-bundle compositions.
All bundle conditions contain the same number of tools; only tool relevance
differs.
}
\label{tab:cross_benchmark_collection}
\end{table}

As shown in Table~\ref{tab:cross_benchmark_collection}, relevant tool
selection substantially affects execution outcomes across all three
benchmarks. Gold bundles consistently achieve higher success rates than
distractor bundles, while half-mixed bundles fall between the two cases in
most settings. These results provide additional evidence that downstream
execution utility depends on query--tool compatibility rather than only the
availability of tools.

\subsection{Cross-Benchmark Utility Prediction}

We next evaluate whether RADEG can predict execution utility across these
independent tool-agent environments. We compare RADEG with a size-only
baseline that uses only the number of tools in the bundle. Since all
conditions contain identical tool counts, the size-only baseline should not
provide meaningful predictive information and serves as a sanity check.

Table~\ref{tab:cross_benchmark_auroc} reports AUROC results across the three
benchmarks. The results show that RADEG consistently outperforms the
size-only baseline, indicating that the gate learns semantic relationships
between queries and available tools rather than relying on superficial bundle
statistics.

\begin{table}[t]
\centering
\small
\setlength{\tabcolsep}{1pt}
\begin{tabular}{lccc}
\toprule
Benchmark
& Size-only
& GoS-based Gate
& RADEG \\
\midrule
$\tau$-Bench (Retail)
& 0.319
& 0.661
& $0.811\pm0.013$ \\

AgentDojo
& 0.348
& 0.456
& $0.687\pm0.033$ \\

ToolSandbox
& 0.478
& 0.555
& $0.952\pm0.014$ \\
\bottomrule
\end{tabular}
\caption{
Cross-benchmark execution utility prediction performance.
Results are AUROC; RADEG results are averaged over five task-level splits
and ten random seeds.
}
\label{tab:cross_benchmark_auroc}
\end{table}

Table~\ref{tab:cross_benchmark_auroc} shows that RADEG maintains strong
utility prediction performance across different tool-agent benchmarks. The
performance variation across datasets reflects differences in task structure,
tool organization, and execution difficulty. In particular, ToolSandbox
exhibits stronger predictability, while AgentDojo contains more heterogeneous
tools and domains, leading to a more challenging prediction setting.

Overall, these additional experiments demonstrate that RADEG is not tied to
the original SkillBench or Graph-of-Skills retrieval pipeline. Instead, the
execution-utility modeling approach remains effective across diverse
tool-agent environments, providing further evidence that execution-aware
gating captures meaningful query--tool compatibility signals.
% \bibliography{aaai2027}

\end{document}